\pdfoutput=1
\documentclass[english]{lni}

\usepackage{epsfig}
\usepackage{graphicx}
\usepackage{subcaption}
\usepackage{amsmath}

\usepackage{amssymb}
\usepackage{todonotes}
\usepackage{siunitx}
\usepackage{booktabs}
\usepackage{multicol}
\usepackage{multirow}
\usepackage{newtxtext,newtxmath}
\usepackage[super]{nth}

\newcommand{\pcent}[1]{\SI{#1}{\percent}}

\begin{document}


\title[Deep Learning Pipeline for Automated Visual Moth Monitoring]{Deep Learning Pipeline for Automated Visual Moth Monitoring: Insect Localization and Species Classification}
\newcommand{\fsu}[1]{Computer Vision Group, Friedrich Schiller University Jena, 07737 Jena, Germany; \email{#1@uni-jena.de}}

\author[
	Dimitri Korsch \and
	Paul Bodesheim \and
	Joachim Denzler
]
{
	Dimitri Korsch
	\footnote{\fsu{dimitri.korsch}}
	\and
	Paul Bodesheim
	\footnote{\fsu{paul.bodesheim}}
	\and
	Joachim Denzler
	\footnote{\fsu{joachim.denzler}}
	\footnote{German Aerospace Center (DLR), Institute for Data Science, Mälzerstraße 3, 07745 Jena, Germany}
	\footnote{Michael Stifel Center Jena for Data-Driven and Simulation Science, Ernst-Abbe-Platz 2, 07743 Jena, Germany}
}


\editor{Gesellschaft für Informatik e.V. (GI)} 

\booktitle{INFORMATIK 2021} 

\year2021


\maketitle


\begin{abstract}
Biodiversity monitoring is crucial for tracking and counteracting adverse trends in population fluctuations.
However, automatic recognition systems are rarely applied so far, and experts evaluate the generated data masses manually.
Especially the support of deep learning methods for visual monitoring is not yet established in biodiversity research, compared to other areas like advertising or entertainment.
In this paper, we present a deep learning pipeline for analyzing images captured by a moth scanner, an automated visual monitoring system of moth species developed within the AMMOD project.
We first localize individuals with a moth detector and afterward determine the species of detected insects with a classifier.
Our detector achieves up to \pcent{99.01} mean average precision and our classifier distinguishes \num{200} moth species with an accuracy of \pcent{93.13} on image cutouts depicting single insects.
Combining both in our pipeline improves the accuracy for species identification in images of the moth scanner from~\pcent{79.62} to~\pcent{88.05}.
\end{abstract}



\section{Introduction} 
\label{sec:introduction}

The discussions and concerns about environmental changes nowadays are both ubiquitous and necessary.
We live in times in which ecosystems change drastically in a short time, and we, as humans, play a significant role in this development.
One of these negative developments is the dramatic loss of insects~\cite{hallmann2017more,wagner2021insect}.
One step towards a better understanding of insect die-off is monitoring species populations, which is time-consuming and often requires experts in that field.
With about one million named species~\cite{stork2018many}, insects represent the vast majority of animals on our planet.
Hence, it is also clear why manual counting of individuals for abundance estimations is not realistic.
Though current developments in big data analysis and computer vision improved a lot, these technologies are not established in insect monitoring as they are, for example, in marketing or entertainment.

Besides others, there are monitoring systems to observe insects~\cite{jonason2014lighttrap,svenningsen2020contrasting,bjerge2021automated}, great apes~\cite{Freytag16_CFW,Brust2017AVM,Kaeding18_ALR,yang2019great,sakib2020visual}, elephants~\cite{Koerschens18:Elephants,Koerschens19:ELPephants}, or sharks~\cite{hughes2017automated}.
Though such monitoring systems are already camera-assisted, a vast amount of data needs to be evaluated.
Unfortunately, the daily work of many ecologists nowadays is still the manual inspection of hundreds or thousands of images, which is exhausting and time-consuming.

As a part of the AMMOD project\footnote{AMMOD = \textbf{A}utomated \textbf{M}ultisensor Station for \textbf{M}onitoring \textbf{o}f Bio\textbf{d}iversity (\url{https://ammod.de/})}, we aim at automated monitoring of species in our immediate vicinity assisted by a computer vision system.
In this paper, we cover the task of categorizing moth species (subset of \emph{Lepidoptera}) by a non-invasive approach.
Within the project, a so-called \emph{moth~scanner} is developed, which consists of an illuminated planar surface and an automated camera system.
During the nighttime, special light sources illuminate the planar area to attract different moth species in the surrounding area.
The automated camera system captures the attracted individuals that land on the illuminated surface.
Finally, our task is to detect and classify the individuals in the taken images.
With the detection and classification results, we assist the ecologists in analyzing the insect population trends.

For automatically analyzing the images, we propose a prototype for a deep learning pipeline consisting of two steps: (1)~localization of individuals via moth detection and (2)~species identification by classification.
For the detection, we use a well-established detection model capable of identifying multiple objects in an image, namely the single-shot MultiBox detector (SSD)~\cite{liu2016ssd}.
The mean average precision (mAP) of our moth detector is~\pcent{88.88} and~\pcent{99.01} for intersection over union (IoU) values above \num{0.75} and \num{0.5}, respectively.

The subsequent classification of \num{200} common moth species in Central Europe is performed with the help of a convolutional neural network (CNN).
In our experiments, we show the benefits of different design decisions for a classifier trained on copped images, namely images depicting a single insect.
The selection of a fine-tuning strategy, the pre-training dataset, and the extension of the classifier with an unsupervised part estimator improve the classification accuracy from~\pcent{63.28} to~\pcent{93.13}.

Finally, we show that the classification accuracy of our proposed pipeline improves with a preceding moth detector on uncropped images.
These are the images captured by the moth scanner, where we cannot ensure that only a single insect has been photographed.
Furthermore, in these images, the insects allocate only a small portion of the entire area.
Hence, preceding a moth classifier with a moth detector, we can improve the classification performance from~\pcent{79.62} to~\pcent{88.05}.


\section{Related Work} 
\label{sec:related_work}

\subsection{Insect Monitoring} 
\label{sub:insect_monitoring}
In general, a commonly used method for monitoring insects is the usage of light traps.
Jonason~\etal~\cite{jonason2014lighttrap} presented a survey on the influence of weather, time of the year, and the type of the light source on the richness and abundance of species.
While the authors identified the moth species manually, some of the first automated species identification systems were presented by Watson~\etal~\cite{watson2004automated}, Mayo and Watson~\cite{mayo2007automatic}, and Batista~\etal~\cite{batista2010classification}.
All of these works used the same dataset, namely \num{35} species with \num{20} individuals per species.
Using support vector machines (SVMs) and nearest neighbor classifiers, they report an accuracy of up to~\pcent{85}~\cite{mayo2007automatic} with leave-one-out cross-validation.
Ding and Taylor~\cite{ding2016automatic} presented automated detection and classification of insect pests.
They used a sliding window approach coupled with a CNN model.
The CNN, followed by a non-maximum suppression as post-processing, performed a binary classification to identify a \emph{codling moth} in the windows.
They achieved an area under the precision-recall curve of \num{0.93}.
In contrast to these works, we perform the classification of much more classes, namely \num{200} moth species.

The works of Chang~\etal~\cite{chang2017fine} and Xia~\etal~\cite{xia2018insect} tackled more challenging classification tasks.
Using images from the Internet, they classified \num{636} and \num{24} species, respectively.
Chang~\etal achieved for the \num{450} butterflies and \num{186} moths species an accuracy of~\pcent{71.5} with a ResNet-18~\cite{resnet} architecture.
Xia~\etal performed a joint detection and classification of individual insects and achieved with their variant of a VGG-19 CNN a mean average precision (mAP) of~\pcent{89.22}.
First, we gather images in a more controlled environment.
As a result, the background is more homogeneous, and the moths are photographed from above in a resting position.
It is worth investigating how far the Internet images that do not represent our desired setup domain may enhance the classification performance.
Anyway, this is out of the scope of this paper.
Furthermore, unlike Xia~\etal, we aim to separate the detection and classification tasks since they will be performed on different physical devices in our setting.

Zhong~\etal~\cite{zhong2018vision} and Bjerge~\etal~\cite{bjerge2021automated} presented detection and classification pipelines deployed on embedded systems, namely on Raspberry Pi variants.
While Zhong~\etal used the YOLO framework~\cite{redmon2016you} for moth detection, Bjerge~\etal presented a detection-by-thresholding approach.
Zhong~\etal achieved a classification accuracy of~\pcent{90.2} for six species with an SVM and shallow features (texture, shape, color, and HOG features).
Bjerge~\etal presented their own CNN architecture and report an F1-score of~\pcent{93.00} for the classification of nine classes.
The authors perform additional counting and tracking of the insects, which is not part of this work.
Furthermore, we outperform the classification results presented by Bjerge~\etal in our experiments (Sect.~\ref{sub:results_on_mcc}).


\subsection{Object Detection} 
\label{sub:related:detection}
Pre-CNN image-based object detection was dominated by Deformable Part Model (DPM)~\cite{felzenszwalb2008dpm} and Selective Search~\cite{uijlings2013selective}.
The first approach uses a sliding window approach, whereas the latter uses region proposal selection as an object detection strategy.
After the rise of CNNs, region proposal methods are dominating the object detection research field.
One of the first was the R-CNN~\cite{girshick2014rcnn} that combined selective search region proposals with a CNN-based classification of these regions.
Many improvement and adaptations based on R-CNN were developed: SPPNet~\cite{he2015sppn}, Fast R-CNN~\cite{girshick2015fast}, MultiBox~\cite{erhan2014multibox}, or Faster R-CNN~\cite{fasterrcnn}.
Some of them improved the classification of the region proposals in quality and computation time~\cite{he2015sppn,girshick2015fast}.
Others improved the quality of the region proposals directly~\cite{erhan2014multibox,fasterrcnn}, especially with an integration of a region proposal CNN.
Some of the methods skip the proposal step and predict bounding boxes directly with the confidences for multiple categories.
Most popular examples are YOLO~\cite{redmon2016you}, OverFeat~\cite{sermanet2013overfeat}, and SSD~\cite{liu2016ssd}.
While OverFeat implements a deep version of the sliding window approach, YOLO uses CNN features to predict bounding boxes and categories.
SSD extracts features from multiple feature maps from multiple stages in the CNN and predicts bounding boxes based on a set of prior locations.

All of them have their advantages and disadvantages.
Meanwhile, there are also dozens of adaptations and improvements to these methods.
Nevertheless, in our work, we use the single-shot MultiBox detector (SSD) since it allows an exchange of the underlying backbone network and yields one of the best results on standard object detection benchmarks like Pascal VOC~\cite{pascal-voc-2007,pascal-voc-2012} and MS COCO~\cite{lin2014mscoco}.


\subsection{Fine-grained Classification} 
\label{sub:related:fgvc}
Fine-grained classification is a special classification task, where the categories, which need to be classified, originate from the same object domain (e.g., bird species~\cite{WahCUB_200_2011}, car models~\cite{StanfordCars}, moth species~\cite{Rodner15:FRD}, or elephant individuals~\cite{Koerschens19:ELPephants}).
The challenge is now to distinguish closely related classes that differ only in subtle features.
In the age of CNNs, it is common to use the data and let the network figure out what are relevant visual features that distinguish a class from the others.
This kind of approach utilizes the input image as it is and performs either smart pre-training strategies~\cite{Cui_2018_CVPR_large,krause2016unreasonable} or advanced feature aggregation techniques~\cite{lin2015bilinear,Simon19:Implicit}.
On the other hand, there are the part- or attention-based approaches~\cite{ge2019weakly,he2019and,zhang2019learning} that extract relevant regions already at the pixel level and use cropped image regions as additional features for the classification.
Both classification strategies have their advantages and drawbacks.
We use an unsupervised approach for part estimation proposed by Korsch~\etal~\cite{Korsch19_CSPARTS} within our pipeline.



\section{Methods} 
\label{sec:methods}

\begin{figure}[t]
	\centering
	\includegraphics[width=\linewidth]{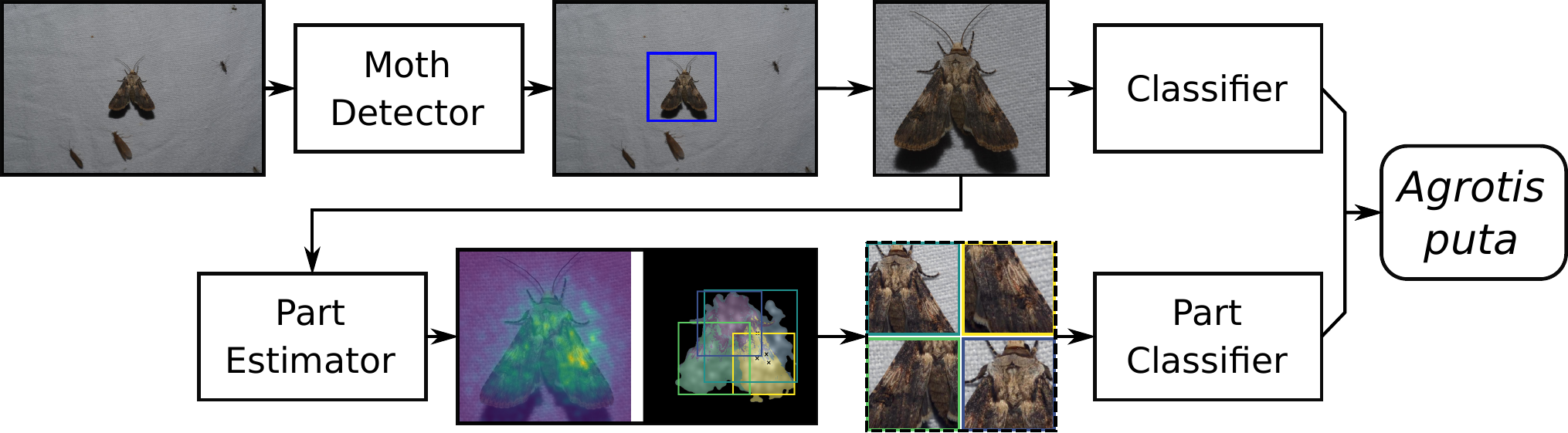}
	\caption{Our deep learning pipeline for automatically analyzing images of the moth scanner: in the first step we use a moth detector to detect an individual in an image. On the one hand, we use this cropped image for the first species prediction. On the other hand, we estimate additional information in form of parts and perform a second part-based prediction. Finally, both predictions are combined to obtain the final species classification.}
	\label{fig:pipeline}
\end{figure}

In this paper, we introduce an automated pipeline for moth species detection and classification.
As visualized in Figure~\ref{fig:pipeline}, the moth detector identifies bounding boxes around the insects given an input image.
Afterward, the image patches identified by the detected bounding boxes are fed into a CNN classifier.
The pipeline performs the classification either only on the input image or extracts informative regions, called parts, which it uses as additional information.
In our experiments, we show that this additional information improves the classification performance (see Sect.~\ref{sub:results_on_eumoths}).

In the following, we introduce the two stages of the pipeline: (1) moth detection based on the single-shot detector (Sect.~\ref{sub:single_shot_detector}), and (2) part-based classification with the help of classification-specific parts (Sect.~\ref{sub:part_based_classification}).

\subsection{Single-shot Detector} 
\label{sub:single_shot_detector}
As already mentioned in Sect.~\ref{sub:related:detection}, the main idea of the single-shot MultiBox detector (SSD) proposed by Liu~\etal~\cite{liu2016ssd} is to utilize feature maps from multiple intermediate stages of the backbone CNN to predict location offsets and class confidences for a set of prior locations.
More precisely, given a feature map with $F \in \mathbb{R}^{N \times M \times P}$ with $P$ channels, $K$ prior bounding boxes with different scales and aspect ratios are defined for each of the $N \cdot M$ locations.
The feature map is transformed by a $3 \times 3 $ convolution with $(C+4) \cdot K$ output channels.
This results for each of the $K$ prior boxes in $C$ per-class scores and four offset values $\Delta=\{dx, dy, dw, dh \}$.
The offsets $dx$ and $dy$ describe the positional offset to the center of the prior box.
The change of the width and the height of a prior box is modeled by $dw$ and $dh$.

\begin{figure}[t]
	\begin{center}
		\begin{subfigure}[t]{.31\textwidth}
			\includegraphics[width=\textwidth]{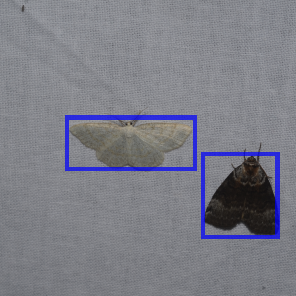}
			\caption{Example image with ground truth annotations.}
			\label{fig:prior_boxes_1}
		\end{subfigure}
		\hfill
		\begin{subfigure}[t]{.31\textwidth}
			\includegraphics[width=\textwidth]{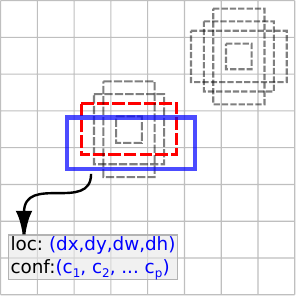}
			\caption{For the first moth a single prior box is selected.}
			\label{fig:prior_boxes_2}
		\end{subfigure}
		\hfill
		\begin{subfigure}[t]{.31\textwidth}
			\includegraphics[width=\textwidth]{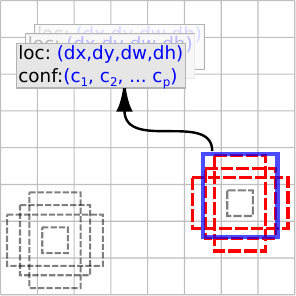}
			\caption{For the second moth multiple prior boxes are selected.}
			\label{fig:prior_boxes_3}
		\end{subfigure}
	\end{center}
	\caption{Example of SSD prior boxes (\emph{dashed gray}) for an $8 \times 8$ feature map. The prior boxes have different scales and aspect ratios, which allows for detecting objects of various sizes and orientations. Different ground truth bounding boxes (\emph{solid blue}) may be matched to a single (\autoref{fig:prior_boxes_2}) or multiple (\autoref{fig:prior_boxes_3}) prior boxes. For each selected prior box (\emph{dashed red}) the offset values and the class scores are estimated. (\emph{Similar to Figure 1 in \cite{liu2016ssd}.)}}
	\label{fig:prior_boxes}
\end{figure}

During training, a prior box is selected as positive when there was a ground truth bounding box with an IoU score higher than~\num{0.5}.
As visualized in Figure~\ref{fig:prior_boxes}, prior boxes at different locations are assigned to different objects.
Additionally, for a single object multiple prior boxes may be selected.
The training objective for estimated location offsets $\Delta$ and class scores $c$ given ground truth bounding boxes $g$ is defined as a sum of the confidence loss and the localization loss:
\begin{equation}
	L(\Delta,c,g) = \dfrac{1}{\mathcal{K}} \left(L_{conf}(c) + L_{loc}(\Delta,g) \right)
\end{equation}
with $\mathcal{K}$ being the number of matched prior boxes, and if no boxes are matched, the loss is set to \num{0}.
The class scores are unnormalized log-likelihoods of a class identified in a certain location.
In our case, the detector distinguishes only the general \emph{moth} class from the background.
For more details about the loss functions, we refer to the original paper of Liu~\etal~\cite{liu2016ssd}.

\subsection{Part-based Classification} 
\label{sub:part_based_classification}

Nowadays, neural networks like CNNs yield the best results in classification by extracting high-level features from the input image in the form of a high-dimensional feature vector (e.g., $D=2048$ in case of InceptionV3).
In the context of a fine-grained recognition task, the classifier has to focus on a specific feature dimension to distinguish a class from the others.
Therefore, we first estimate the most informative features for the current classification task.
It is realized by utilizing a linear classifier with a sparsity-inducing L1-regularization.
An optimization with L1-regularization forces the classifier's decisions to perform the classification on a small subset of feature dimensions.
This kind of implicit feature selection is classification-specific.
Furthermore, it allows identifying for each class the feature dimensions that best distinguish this class from all other classes by selecting dimensions with classifier weights above a certain threshold.

\noindent\textbf{Informative Image Regions:}\quad
We utilize gradient maps~\cite{simonyan2013deep} to estimate the most informative pixels in the image, identified by large gradients.
As described previously, we restrict the computation of the gradients only to the feature dimensions used by the L1-regularized classifier.
Thus, we incorporate the initial classification in the estimation of the part regions.
Like Simonyan \etal~\cite{simonyan2013deep} and Simon \etal~\cite{Simon_2015_ICCV}, we use back-propagation through the CNN to identify the regions of interest for each selected feature dimension.
We compute a saliency map $\vec{M}(\vec{I})$ for an image $\vec{I}$ based on the feature dimension subset $\mathfrak{D} \subset \{1, \dots, D\}$ as follows:

\begin{equation}
	\label{eq:saliency_sum}
	M_{x,y}(\vec{I})
	  = \dfrac{1}{|\mathfrak{D}|} \sum_{d \in \mathfrak{D}} \left| \dfrac{\partial}{\partial I_{x,y}} f^{(d)}(\vec{I}) \right|
	  \quad  .
\end{equation}

\noindent\textbf{Part Estimation:}\quad
Next, we normalize the values of the saliency map to the range $[0 \dots 1]$, and discard regions of low saliency by setting values beneath the mean saliency value to $0$.
We use the resulting sparse saliency map to estimate the spatial extent of coherent regions.
Like Zhang \etal~\cite{zhang2019unsupervised}, we achieve this by \mbox{$k$-means} clustering of pixel coordinates $(x, y)$ and the saliencies $M_{x,y}$ (Eq.~\ref{eq:saliency_sum}).
Additionally, we also consider the RGB values at the corresponding positions in the input image.
The clusters are initialized with $k$ peaks computed by non-maximum suppression, identifying locations with the largest saliencies.
Consequently, the number of peaks determines the number of parts to detect.
Finally, it is straightforward to identify a bounding box around each estimated cluster, and the resulting bounding boxes serve as parts for the following classification.

\noindent\textbf{Extraction and Aggregation of Part Features:}\quad
In the final step, we extract image patches with the help of the estimated bounding boxes and treat them as regular images.
The neural network should extract different features from these image patches than from the original image because the level of detail varies between these types of input.
Therefore, we process the part images by the same CNN architecture as the original image but with a separate set of weights.
Afterward, for every part image, the CNN extracts a feature vector, resulting in a set of part features for every single image.
There are different ways to aggregate these features to a single feature vector and perform the classification.
We have chosen to average over the part features, which results in a single feature vector with the same dimension as for the original image.
This aggregation strategy yielded better results in our experiments than, for example, concatenation of part features.
Finally, classification is performed based on the global feature and the aggregated part feature.
For joint optimization of both CNNs, we average the cross-entropy losses of the global prediction $\vec{p}$ and part prediction $\vec{q}$.
It equals to computing the geometrical mean of normalized class probabilities and enforces both classifiers to be certain about the correct class:

\begin{align}
	L_{final}\left(\left\{\vec{p},\vec{q}\right\}, y\right)
		&= \frac{1}{2}\left(L\left(\vec{p}, y\right) + L\left(\vec{q}, y\right) \right) \\
		&= -\frac{1}{2}\left(
			\sum^{C}_{i=1} y_i \cdot \log (p_{i}) +
			\sum^{C}_{i=1} y_i \cdot \log (q_{i}) \right)\\
		&= -\sum^{C}_{i=1} y_i \cdot \log \left(
			\sqrt{p_i \cdot q_i}
		\right) \quad .
\end{align}




\section{Experiments}
\label{sec:experiments}

In the following, we evaluate the two parts of our moth scanner pipeline.
We perform each experiment ten times and provide in Sections~\ref{sub:results_on_mcc}, and~\ref{sub:results_on_eumoths} the mean and the standard deviation of the evaluation metrics across the different runs.
We fine-tune all CNNs for \num{60} epochs with the RMSProp~\cite{tieleman2012lecture} optimizer and L2-regularization with a weight decay of~\num{5e-4}.
The starting learning rate of \num{1e-4} is reduced by \num{0.1} after \num{20} and \num{40} epochs.
Furthermore, we utilize standard image augmentation methods: random cropping, random horizontal and vertical flipping, and color jittering (contrast, brightness, and saturation).
In the case of the classification, we further utilize label smoothing regularization~\cite{inception} with a smoothing factor of \num{0.1}.

First, we evaluate the performance of the moth detector presented in Sect.~\ref{sub:single_shot_detector}.
In Table~\ref{tab:detection_results}, we report the mean average precision (mAP) as the evaluation metric for the detections.
The precision is based on predictions of detected objects, where the intersection over union (IoU) of predicted and ground truth bounding boxes is above a certain threshold.
The IoU describes how well two bounding boxes match by computing the ratio between the intersection and the union of the areas of the two bounding boxes.
The two typical evaluation metrics used in one of the most common object detection benchmarks~\cite{lin2014mscoco} are \emph{mAP@0.5} and \emph{mAP@0.75}, with IoU thresholds of \num{0.5} and \num{0.75}, respectively.
We use a standard version of the detection network with an input size of $300\times300$ and the VGG16~\cite{simonyan2014very} architecture as a backbone that is pre-trained on ImageNet~\cite{ImageNet}.
All additional layers included for the detection task and not initially present in the VGG architecture are initialized randomly.

Second, we evaluate the classification performance.
Since the datasets we use have a balanced sample distribution across the classes, we use accuracy as a primary evaluation metric.
To be independent of the underlying detector, we use the ground truth bounding boxes and perform the classification on the cropped image patches.
Further, we extract additional parts, as described in Sect.~\ref{sub:part_based_classification}, and combine the predictions on these parts with the predictions on the entire image.
For the classification, we use the InceptionV3 CNN architecture~\cite{inception}.
Here, we also decided on the default input size of $299\times299$ for both the images and parts.
Furthermore, we investigate the effect of two different pre-training methods.
As the typical choice, we use ImageNet~\cite{ImageNet} pre-training since most of the deep learning frameworks implementing different CNN architectures also provide weights for these architectures pre-trained on the ImageNet dataset.
Additionally, we use pre-training on the iNaturalist~\cite{iNaturalist} dataset provided by Cui~\etal~\cite{Cui_2018_CVPR_large}.
Data used in this kind of pre-training is more related to the domain of animals, in our case to the domain of insects, which can also be seen in the improvement of the classification accuracy in Table~\ref{tab:classification_results:eumoths}.

Finally, we evaluate the proposed pipeline as a whole: given an uncropped image, like in Figure~\ref{fig:examples:eumoths}, we first apply the moth detector and then perform the classification on the resulting image patches.
We compare this setup to a classifier that performs the classification on the initial uncropped images.

We further evaluate the classifier on the dataset provided by Bjerge~\etal~\cite{bjerge2021automated}.
Unfortunately, the authors report only F1-scores of their classification and do not provide their training-test split.
Nevertheless, we performed five-fold cross-validation with the same train-test ratios as in the original paper.
For each of the folds, we repeated the training ten times, like in the previous experiments.

\subsection{Datasets}
\label{sub:datasets}


\begin{figure}[t]
	\centering
	\begin{subfigure}[t]{0.49\textwidth}
		\includegraphics[width=\linewidth]{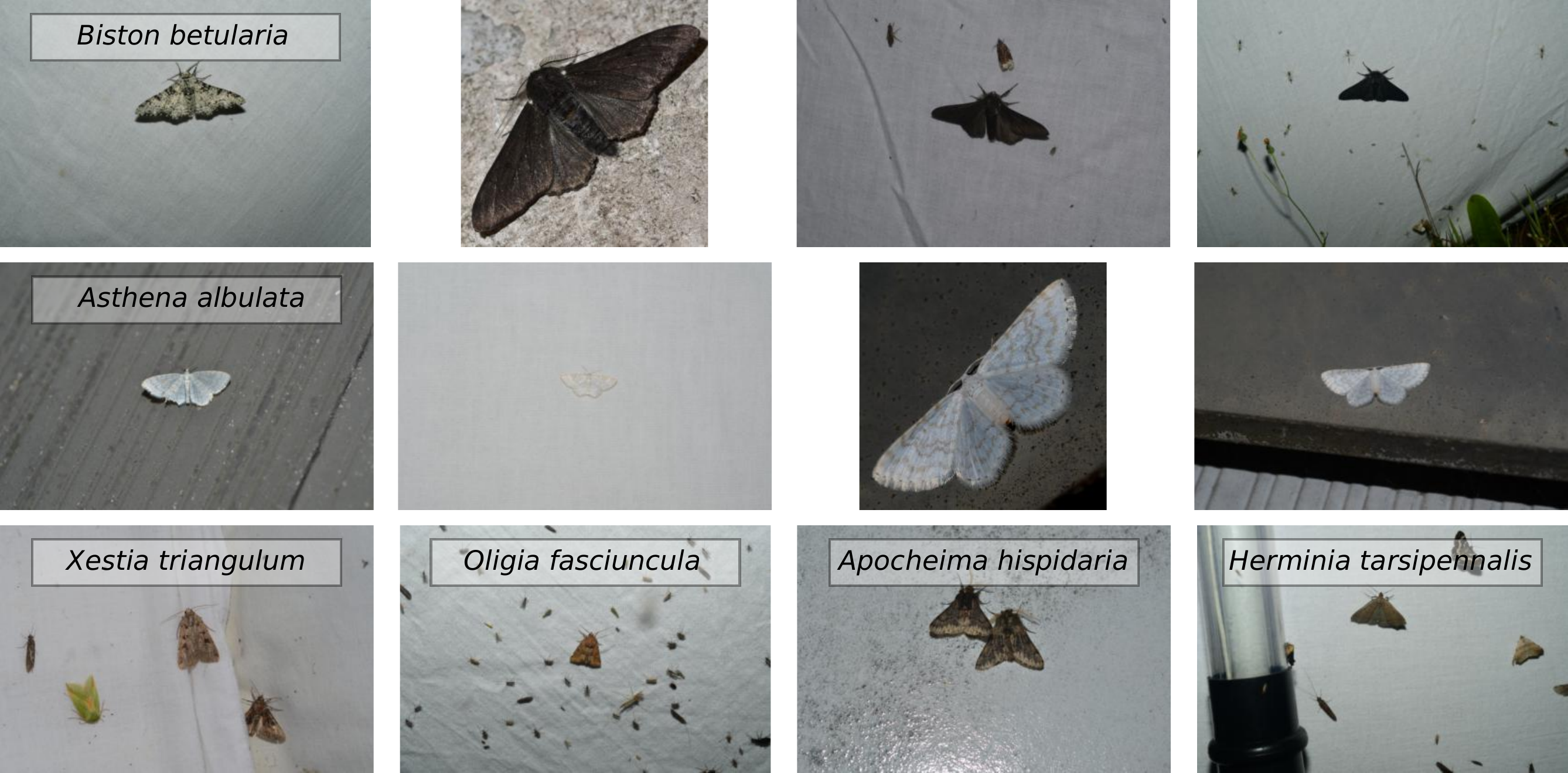}
		\caption{The first two rows show two different moths species, whereas the third row shows images with more than one insect. These examples illustrate the versatility in the appearance of the moths.}
		\label{fig:examples:eumoths}
	\end{subfigure}
	\hfill
	\begin{subfigure}[t]{0.49\textwidth}
		\includegraphics[width=\linewidth]{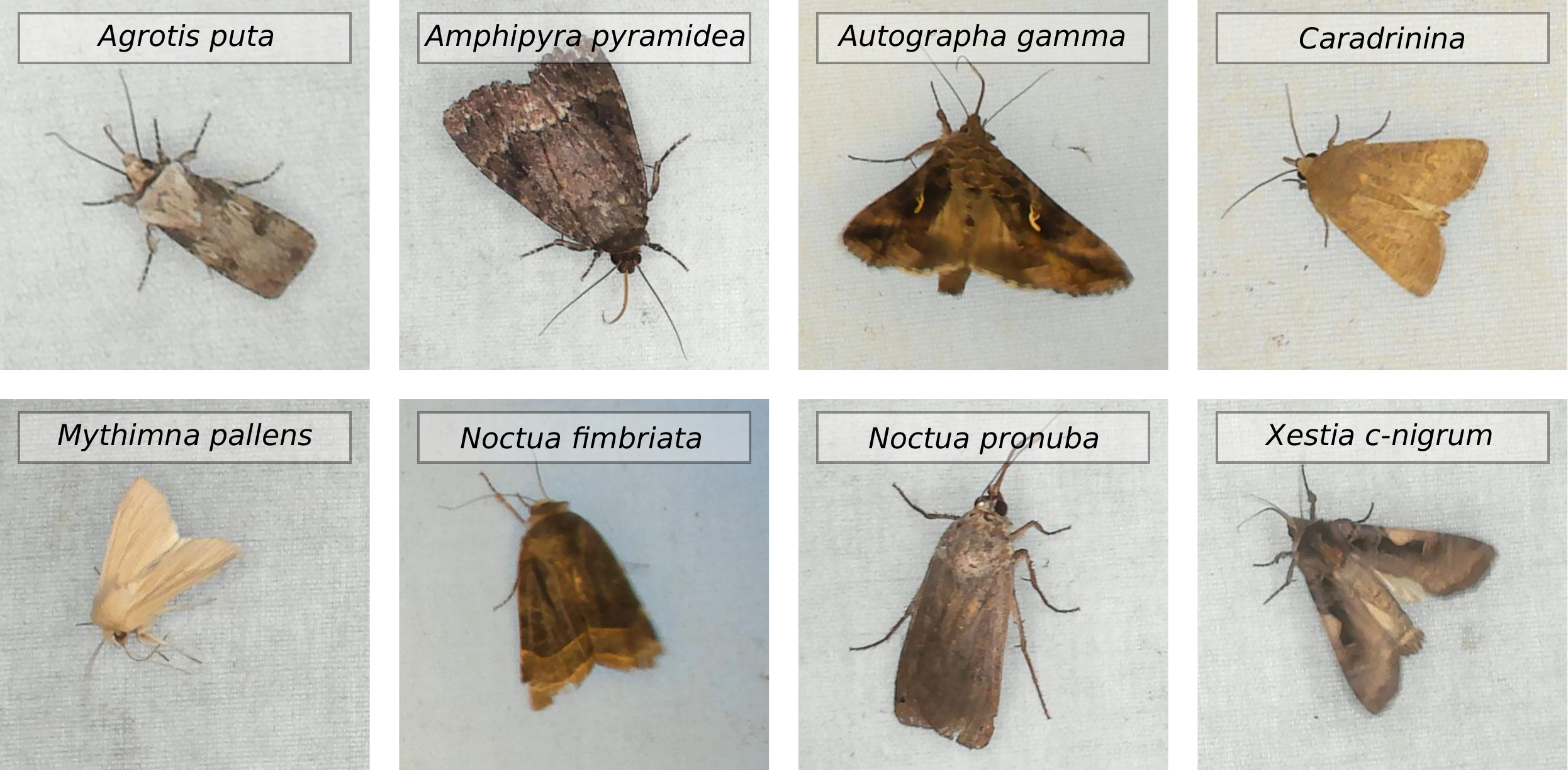}
		\caption{Cropped images of the eight MCC classes. Bjerge~\etal~\cite{bjerge2021automated} used an eight-megapixel web camera to capture the images. Hence, the details of the species are barely visible.}
		\label{fig:examples:mcc}
	\end{subfigure}
	\caption{Example images from the \emph{EU-Moths} (a) and \emph{MCC} (b) datasets.}
	\label{fig:examples}
\end{figure}

\noindent\textbf{Moth Classification and Counting (MCC) Dataset\footnote{\url{https://github.com/kimbjerge/MCC-trap}}:}\quad Created by Bjerge~\etal~\cite{bjerge2021automated}, the subset for the classification consists of eight moth species with \num{250} images for each species, resulting in a dataset of \num{2000} images.
The authors used an 80:20 training-test split of the data but did not provide their specific split.
Additionally, there are two more classes: a background class and a class for a wasp species.
We ignore these classes and perform the training and the evaluation only on the eight moths species.
An individual from every class is shown in Figure~\ref{fig:examples:mcc}.
Compared to the EU-Moths dataset, the images are of lower quality since the authors captured them with an eight-megapixel web camera connected to a Raspberry Pi 4.

\noindent\textbf{European Moths (EU-Moths) Dataset\footnote{\url{https://www.inf-cv.uni-jena.de/eu_moths_dataset}}:}\quad From roughly \num{4700} moth species present in Central Europe\footnote{\url{http://lepiforum.org/} (accessed on \nth{6} July, 2021)}, this dataset consists of \num{200} species most common in this region.
Each of the species is represented by approximately \num{11} images.
We considered a random but balanced training-test split in eight training and three test images per species, resulting in roughly \num{1600} training and \num{600} test images in total.
To evaluate the detector, we manually annotated the bounding boxes around the insects.
Some examples of the images are shown in Figure~\ref{fig:examples:eumoths}.
The insects are photographed manually and mainly on a relatively homogeneous background.
About \pcent{92} of the images contain only a single individual like it is shown in the first two rows of Figure~\ref{fig:examples:eumoths}.
The last row of the same image depicts images with more than one insect of interest.
This scenario may happen in the final moth scanner installation, and it is crucial to test how the detector performs in this case.

This dataset yields different challenges for the detector and the classifier.
On the one hand, the detector should be able to detect insects of different sizes.
Furthermore, we require a detector with MultiBox support.
Both of these properties are fulfilled by the SSD.
On the other hand, designing a classifier that can predict species from these raw images is difficult.
As proposed in this paper, we decided to use a moth detector to locate single insects and classify these separately.
Hence, the classifier is trained on images cropped to the bounding boxes identifying a single individual.


\begin{figure}[t]
	\begin{center}
	\begin{subfigure}[t]{0.525\textwidth}
		\includegraphics[width=\linewidth]{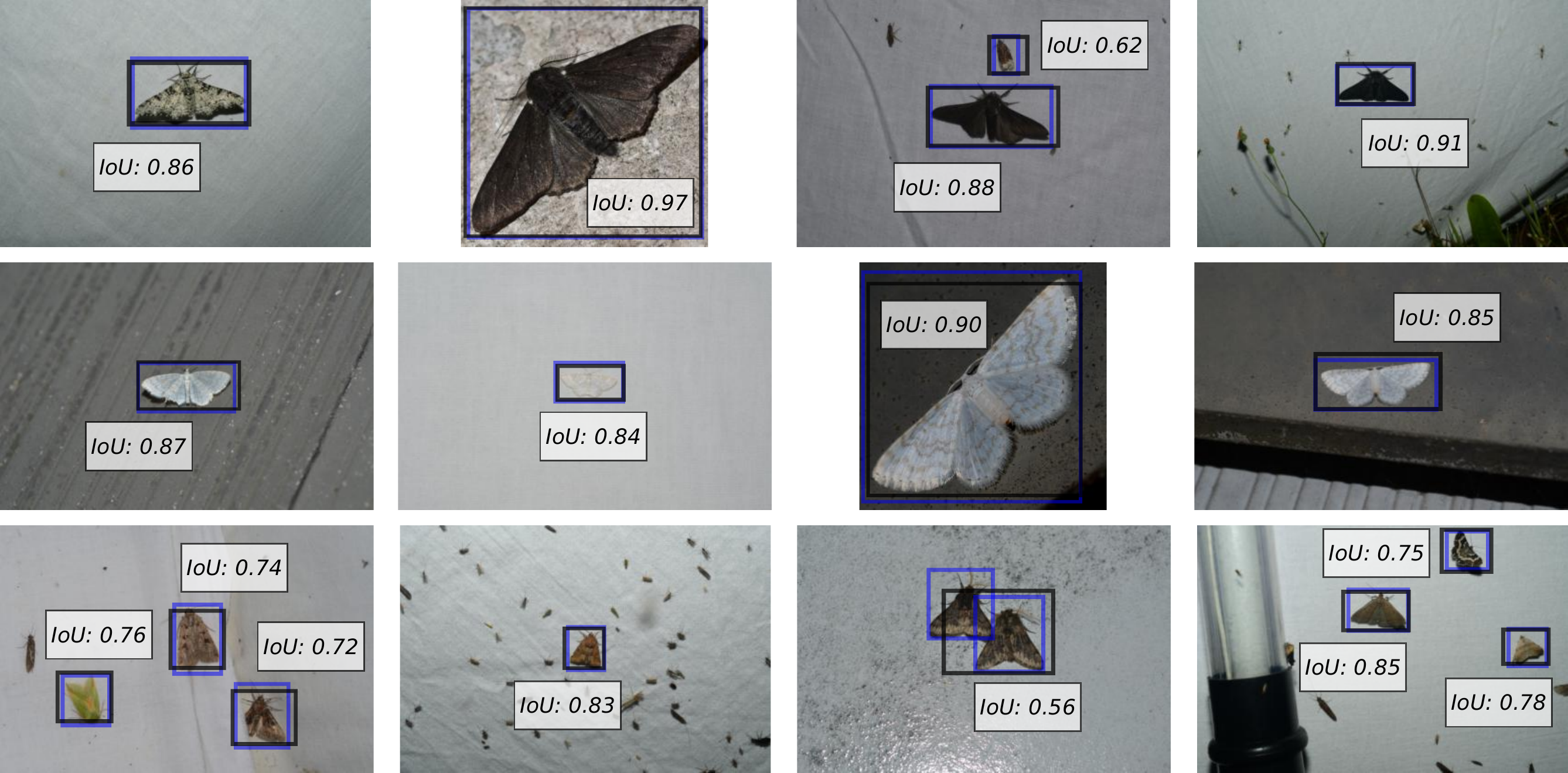}
		\caption{Our manual ground truth annotations are visualized in \emph{blue}. Text boxes contain the IoU of each detection with the ground truth.}
		\label{fig:example_detections:eumoth}
	\end{subfigure}
	\hfill
	\begin{subfigure}[t]{0.46\textwidth}
		\includegraphics[width=\linewidth]{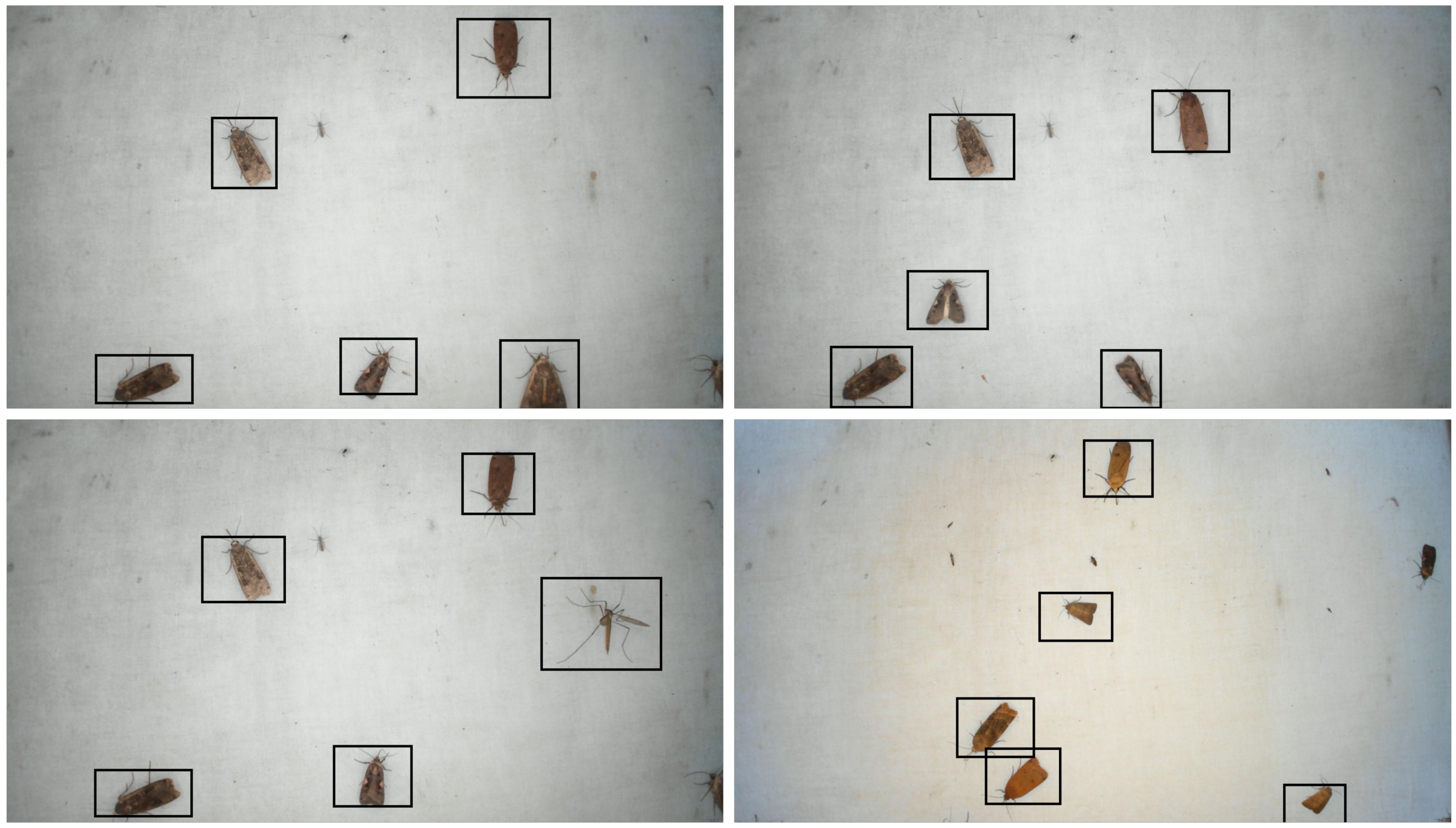}
		\caption{The MCC dataset does not provide any ground truth annotations. Hence, only the detections are visualized.}
		\label{fig:example_detections:mcc}
	\end{subfigure}

	\caption{Visualized detection results (\emph{black} bounding boxes) on the \emph{EU-Moths} (a) and \emph{MCC} (b) datasets.}
	\label{fig:example_detections}
	\end{center}
\end{figure}

\subsection{Results on the MCC Dataset} 
\label{sub:results_on_mcc}


\begin{table}[t]
	\begin{center}
	\begin{tabular}{lccc}
		\toprule
		 & \multirow{2}{*}{\textsc{F1-Score}}
		 &  \multicolumn{2}{c}{\textsc{Accuracy}} \\

		 &
		 & \textsc{ImageNet} & \textsc{iNaturalist} \\
		\midrule
		\textsc{Bjerge~\etal~\cite{bjerge2021automated}}
			& \pcent{93.00}
			& --
			& -- \\
		\textsc{Ours}
			& \pcent{99.69} \scriptsize{$(\pm \pcent{0.34})$}
			& \pcent{99.41} \scriptsize{$(\pm \pcent{0.79})$}
			& \pcent{99.55} \scriptsize{$(\pm \pcent{0.40})$} \\
		\bottomrule
	\end{tabular}
	\caption{Comparison of the classification results on cropped images (see Figure~\ref{fig:examples:mcc}) of the MCC dataset provided by Bjerge~\etal~\cite{bjerge2021automated}. Besides the F1-Score that was reported by the authors in their work, we report additionally the accuracy for our trained classifiers.}
	\label{tab:classification_results:mcc}
	\end{center}
\end{table}

\paragraph{Detection:} Unfortunately, the authors do not provide any bounding box annotations.
Hence, we were not able to evaluate the performance of our moth detector on this dataset.
Nevertheless, we provide a qualitative evaluation on some of the images in Figure~\ref{fig:example_detections:mcc}.

\paragraph{Classification:}
In Table~\ref{tab:classification_results:mcc}, we compare  our classification method with the one proposed by Bjerge~\etal~\cite{bjerge2021automated} on the MCC dataset.
One can see that our classifier achieves near-perfect accuracies and F1-scores.
We assume the reason for these results is in the composition of the dataset.
Since Bjerge~\etal do not provide any training-test split, we have used a random split with the same ratio (80:20) as the authors in their work.
Nevertheless, one can see in Figure~\ref{fig:example_detections:mcc} some of the moth individuals do not move in different images captured in short intervals.
As a result, extracting image crops from these images would result in near-identical images in different splits after the random splitting.
Consequently, one would train and test the classifier on nearly the same data.
Nevertheless, to be comparable to the results of Bjerge~\etal, we chose the same splitting strategy, even though it may not represent the correct evaluation of the model.

\subsection{Results on the EU-Moths Dataset} 
\label{sub:results_on_eumoths}


\begin{table}[t]
	\begin{center}
	\begin{tabular}{lcc}
		\toprule
		 & \textsc{mAP@0.75} & \textsc{mAP@0.5} \\
		\midrule
		\textsc{Evaluated on}
			& \multicolumn{2}{c}{\textsc{Trained on entire dataset}} \\
		\textsc{Entire dataset}
			& \pcent{88.88} \scriptsize{$(\pm \pcent{0.77})$}
			& \pcent{99.01} \scriptsize{$(\pm \pcent{0.09})$} \\
		\midrule
		\textsc{Evaluated on}
			& \multicolumn{2}{c}{\textsc{Trained on subset 1}} \\
		\textsc{Subset 1}
			& \pcent{87.38} \scriptsize{$(\pm \pcent{1.65})$}
			& \pcent{98.53} \scriptsize{$(\pm \pcent{0.13})$} \\
		\textsc{Subset 2}
			& \pcent{78.83} \scriptsize{$(\pm \pcent{1.11})$}
			& \pcent{99.19} \scriptsize{$(\pm \pcent{0.28})$} \\
		\midrule
		\textsc{Evaluated on}
			& \multicolumn{2}{c}{\textsc{Trained on subset 2}} \\
		\textsc{Subset 1}
			& \pcent{85.88} \scriptsize{$(\pm \pcent{1.04})$}
			& \pcent{99.70} \scriptsize{$(\pm \pcent{0.10})$} \\
		\textsc{Subset 2}
			& \pcent{82.45} \scriptsize{$(\pm \pcent{1.23})$}
			& \pcent{98.04} \scriptsize{$(\pm \pcent{0.29})$} \\
		\bottomrule
	\end{tabular}
	\caption{Detection results on the EU-Moths dataset. First row contains the evaluation of the detectors trained on the entire dataset (\num{200} classes). The lower part of the table shows the capability of the detector to localize unseen species. For that, we split the dataset in two parts with distinct classes (\textsc{subset 1}: classes \SIrange{1}{100}{}, and \textsc{subset 2}: classes \SIrange{101}{200}{}) and perform cross-subset evaluations.}
	\label{tab:detection_results}
	\end{center}
\end{table}

\paragraph{Detection:}
We split this experiment into two parts: (1) we evaluated the detection performance on the entire dataset, and (2) we split the dataset into two subsets of distinct classes.
The first part evaluates the standard performance of the detector.
The second part of the experiment evaluates how good the detector performs on classes not seen during the training.
This scenario is essential since not all species may be available at training time in the real-world setup.
One could train the detector on a dataset captured at one location and deploy it at another one.
Furthermore, a detector able to localize moth species not seen at training time is beneficial for novelty detection, active learning, and incremental learning algorithms~\cite{Brust2017AVM,Brust2020AIL}.

Table~\ref{tab:detection_results} shows the detection results for both experiments.
As previously mentioned, we report the mean	average precision for IoU thresholds 0.75 and 0.5 (mAP@0.75 and mAP@0.5).
The first row shows the results for the first experiment.
The detector seems to be quite precise if we consider the challenges of the dataset.
The lower part of Table~\ref{tab:detection_results} further shows the cross-subset results.
Here we can see that the mAP@0.75 performance drops compared to the previous experiment, and the standard deviation increases.
Both are explainable because, for the second experiment, we used only half of the classes for the training.
Furthermore, mAP@0.5 performance remains comparable to the first experiment, which shows the moth detector's reliability for unseen classes.

Additionally, Figure~\ref{fig:example_detections} depicts qualitative results of the detector.
In Figure~\ref{fig:example_detections:eumoth}, we estimated the bounding boxes (in \emph{black}) for some of the images of the EU-Moths dataset.
We visualized the ground truth bounding boxes (in \emph{blue}) and the resulting IoU between the prediction and the ground truth.
The detector's most significant challenge seems to be insects located too close to each other (second last example in the final row).


\begin{table}[t]
	\centering
	\begin{tabular}{llcc}
		\toprule

		 & & \multicolumn{2}{c}{\textsc{Accuracy}} \\

		 & \textsc{Fine-tuning} & \textsc{ImageNet} & \textsc{iNaturalist} \\
		\midrule
		\multirow{2}{*}{\textsc{No Parts}}
			& \emph{only FC layer}
			& \pcent{63.28} \scriptsize{$(\pm \pcent{0.45})$}
			& \pcent{86.60} \scriptsize{$(\pm \pcent{0.42})$} \\

			& \emph{entire CNN}
			& \pcent{89.46} \scriptsize{$(\pm \pcent{0.88})$}
			& \pcent{90.54} \scriptsize{$(\pm \pcent{1.10})$} \\
		\midrule
		\multirow{2}{*}{\textsc{With Parts}}
			& \emph{only FC layer}
			& \pcent{71.82} \scriptsize{$(\pm \pcent{0.35})$}
			& \pcent{87.96} \scriptsize{$(\pm \pcent{0.38})$} \\
			& \emph{entire CNN}
			& \pcent{91.50} \scriptsize{$(\pm \pcent{0.61})$}
			& \pcent{93.13} \scriptsize{$(\pm \pcent{0.76})$} \\
		\bottomrule
	\end{tabular}
	\caption{Classification results on cropped images of the EU-Moths dataset. The results show the effects of the different fine-tuning strategies, the two pre-training datasets, and the usage of additional information in the form of parts.}
	\label{tab:classification_results:eumoths}
\end{table}

\paragraph{Classification:}
Table~\ref{tab:classification_results:eumoths} shows the results of the classification.
We compare different training and pre-training methods and whether the additional information in the form of parts benefits the classification accuracy.

First, one can see that fine-tuning the entire CNN instead of using it only as feature extractor results in an improvement of the recognition rate by roughly \pcent{26} and \pcent{4} for CNNs pre-trained on ImageNet~\cite{ImageNet} and iNaturalist 2017~\cite{iNaturalist} datasets, respectively.
Though training the entire CNN results in longer training times and is computationally more expensive, the improvements are visible.

Second, the choice of the pre-training dataset is also crucial.
Replacing the typical CNN weights provided by almost every deep learning framework pre-trained on the ImageNet dataset with the ones proposed by Cui~\etal~\cite{Cui_2018_CVPR_large} leads to a further improvement.
The later pre-training increases the accuracy by approximately \pcent{1} if the entire CNN is trained.
It also yields remarkable benefits if choosing computationally cheaper training of only the final classification layer, namely an improvement of \pcent{20}.

Finally, employing additional information in the form of parts improves the classification accuracies by approximately \SIrange[range-units=single]{2}{2.6}{\%} depending on the chosen pre-training.
We achieved the best results with the part-based setup: \pcent{91.50} and \pcent{93.13} with ImagenNet and iNaturalist pre-training, respectively.

\paragraph{Entire Pipeline:}

\begin{table}[t]
	\centering
	\begin{tabular}{lc}
		\toprule
		\textsc{Composition} & \textsc{Accuracy} \\
		\midrule
		\textsc{Classifier Only}
			& \pcent{79.62} \scriptsize{$(\pm \pcent{1.10})$} \\
		\textsc{Detector + Classifier}
			& \pcent{88.05} \scriptsize{$(\pm \pcent{0.58})$} \\
		\bottomrule
	\end{tabular}
	\caption{Classification results on uncropped images as shown in Figure~\ref{fig:examples:eumoths}. The effect of a preceding detector on the classification accuracy is clearly visible.}
	\label{tab:pipeline_results}
\end{table}

In this experiment, we evaluate the entire pipeline as presented in Sect.~\ref{sec:methods}.
For this purpose, we couple every trained detector with every trained classifier and observe the resulting classification performance.
This way, we report the mean accuracy of \num{100} detector-classifier combinations in the last row of Table~\ref{tab:pipeline_results}.
As a baseline method, we trained ten CNN classifiers on the original uncropped images.
One can see that the preceding detector improves the classification accuracy by approximately \pcent{9}.


\section{Conclusions}
\label{sec:conclusions}

In this paper, we presented an automated detection and classification pipeline for \num{200} moth species.
We plan to deploy this pipeline in the visual monitoring system of the AMMOD project, the so-called moth scanner, which will help ecologists observe the population trends of the insects.
Since light sources easily attract moths, the moth scanner consists of an illuminated white surface and a camera that automatically captures images of insects resting on this surface.
We first localized the moths with a single-shot MultiBox detector (SSD) in the recorded images and then classified the resulting detections using a CNN classifier.
We also showed the effect of different training configurations on the final classification accuracy: the choice of fine-tuning strategy, the selection of the pre-training dataset, and the extension of the classification with an unsupervised part estimator.
In our experiments, each part of the pipeline achieved promising results: a detection rate of up to \pcent{99.01} (mAP@50) and classification accuracy on images depicting a single insect of up to \pcent{93.13}.
Finally, we evaluated both parts of the pipeline together and improved the classification accuracy on original images captured by the moth scanner from~\pcent{79.62} to~\pcent{88.05} compared to a setup without a preceding moth detector.

\section*{Acknowledgments}
This work has been funded by the German Federal Ministry of Education and Research (Bundesministerium für Bildung und Forschung, BMBF, Deutschland) via the project ``\mbox{Development} of an \mbox{Automated} \mbox{Multisensor} \mbox{Station} for \mbox{Monitoring} of \mbox{Biodiversity} (\mbox{AMMOD}) - Subproject 5: Automated Visual Monitoring and Analysis'' (FKZ: 01LC1903E).

\bibliography{main}
\end{document}